\documentclass[]{llncs}
\usepackage{graphicx}
\usepackage[left=2cm, right=2cm, top=2cm]{geometry}
\usepackage{txfonts}
\usepackage{pxfonts}
\usepackage{csquotes}
\usepackage{setspace}

\usepackage[citestyle=authoryear,natbib=true,backend=bibtex]{biblatex}
\bibliography{bib}

\begin{document}
\title{A General Counterexample to Any Decision Theory and Some Responses}

\author{Joar Skalse}

\institute{Oxford University \newline
\email{joar.skalse@philosophy.ox.ac.uk}}



%
\maketitle

\begin{abstract}
In this paper I present an argument and a general schema which can be used to construct a problem case for any decision theory, in a way that could be taken to show that one cannot formulate a decision theory that is never outperformed by any other decision theory. I also present and discuss a number of possible responses to this argument. One of these responses raises the question of what it means for two decision problems to be \enquote{equivalent} in the relevant sense, and gives an answer to this question which would invalidate the first argument. However, this position would have further consequences for how we compare different decision theories in decision problems already discussed in the literature (including e.g.\ Newcomb's problem).
\end{abstract}


Suppose there exists a decision theory XDT such that an agent who follows the recommendations of XDT is at least as well-off as any other agent in any \enquote{fair} decision problem. I will present an argument for why no such decision theory XDT can exist. Let DP(XDT) be the following decision problem:
\begin{quote}
    There are two boxes in front of you. One box is transparent and can be observed to contain \$1,000, and one box is opaque. You can choose to take either of the two boxes, but not both. The boxes have been set up in such a way that if XDT permits taking the opaque box then the opaque box is empty, else the opaque box contains \$2,000. The decision maker knows this fact.
\end{quote}
I will argue that whatever XDT recommends in DP(XDT), we can construct an alternative decision theory YDT such that followers of YDT are better off than followers of XDT in DP(XDT). I will, for the time being, assume that DP(XDT) is a \enquote{fair} decision problem in the relevant sense. I will also assume that a satisfactory decision theory always must permit at least one action in any decision problem.\footnote{It is worth noting that this would mean that e.g.\ ratificationism (\cite{Jeffrey1965}) is not a \enquote{satisfactory} decision theory, since there are decision problems in which there are no ratifiable choices (see e.g.\ \cite{Egan2007}). However, refinements such as \textit{lexical ratificationism} (\cite{Egan2007}) do satisfy this condition.} This means that we have three options; either XDT only permits taking the transparent box, or XDT only permits taking the opaque box, or XDT permits taking either the transparent box or the opaque box. In the first two cases we could simply let YDT invert the recommendation of XDT. In this case, a decision maker who follows the recommendation of XDT will end up with \$1,000 less than a decision maker who instead follows the recommendation of YDT, which contradicts the assumption that followers of XDT are at least as well-off as all other agents in all \enquote{fair} decision problems.

What if XDT permits taking either of the two boxes? Formally, there is then some set of probabilities $S \subseteq [0,1]$ such that, for each $p \in S$, XDT permits taking the opaque box with probability $p$, and where $S \neq \{1\}$ and $S \neq \{0\}$. In this case, the opaque box will be empty (since XDT permits taking the opaque box). Let YDT recommend taking the transparent box (with probability $1$). Let $\mathcal{X}$ be an agent that follows XDT and in DP(XDT) takes the transparent box with probability $p \in S$, $p \neq 1$, and let $\mathcal{Y}$ be an agent that follows YDT. In this case, $\mathcal{Y}$ is guaranteed to get at least as much money as $\mathcal{X}$, and might get more money than $\mathcal{X}$, which again contradicts the assumption that followers of XDT always are at least as well-off as any other agent in any \enquote{fair} decision problem. This exhausts all the available options, and so it seems like no such decision theory XDT can exist.

Before moving on, I want to make a few clarifying remarks. What I have presented above is a general schema which can be used to construct a seemingly problematic case for \textit{any} decision theory. For example, we can consider what YDT does in DP(YDT), or what causal decision theory (CDT) and evidential decision theory (EDT) do in DP(CDT) and DP(EDT) respectively. If the recommendation of a given decision theory ZDT depends on the Bayesian prior $\mathcal{D}$ of the decision maker then we could simply modify the schema to take the prior into account as well -- i.e., we construct a decision problem DP(ZDT, $\mathcal{D}$) where the opaque box contains \$2,000 if ZDT does not permit agents with the prior $\mathcal{D}$ to take the opaque box. Therefore, whatever the structure of a given decision theory is, this schema can be used to construct a problem case for that decision theory. I should also comment on the case where XDT permits taking either of the two boxes, but allows the agent to take the opaque box with probability $p=1$ (i.e., $1 \in S$). In that case the opaque box will contain \$2,000, and an agent $\mathcal{A}$ could follow XDT and take the opaque box with probability $1$ in DP(XDT), in which case no agent is better off than $\mathcal{A}$ in DP(XDT). However, in this case an XDT agent is still \textit{allowed} to not take the opaque box with probability $1$, so \textit{merely} following XDT is not \textit{sufficient} to ensure that you are at least as well-off as any other agent in any decision problem. And, of course, if we construct a refined decision theory XDT$^+$ which says \enquote{follow XDT, and if you face DP(XDT) then take the opaque box with probability $1$}, then we could use the schema to construct a decision problem PD(XDT$^+$), which would be problematic for this refined decision theory.

One possible response to this argument is to say that DP(XDT) in some sense is \enquote{unfair}, and that we should exclude it from the set of situations we consider. However, we must then provide some principle on the grounds of which DP(XDT) can be excluded, and it is not immediately obvious how to do this. We can note that it is not the case that DP(XDT) is an \enquote{unwinnable} decision problem. In DP(XDT) there is a choice that is available to the decision maker that is better than the choice that XDT in fact recommends, and an agent that follows YDT will actually get more money than an XDT agent. Moreover, it is also not the case that DP(XDT) rewards or punishes decision makers based directly on what decision theory they follow -- it is possible to set up DP(XDT) without knowing what decision theory the decision maker follows, and the outcome depends only on what action the decision maker takes. This means that DP(XDT) is different from \enquote{decision} problems in which the decision maker is directly punished for following some particular decision theory, for example.

We should also note that this decision problem is different from Newcomb's problem (\cite{Nozick1969}) in a number of important ways. In particular, many of the objections that are sometimes raised against Newcomb's problem do not apply here. A causal decision theorist can argue that an evidential decision theorist and a causal decision theorist do not actually have the same options available to them when they are faced with Newcomb's problem (see e.g.\ pp.\ 151--154 in \cite{Joyce1999-JOYTFO-4}). The reason that evidential decision theorists seem to do better is because they are given better options, but given the options that they have the decision that they make is irrational (or so one might argue). However, this argument does not apply to DP(XDT), since both the state of the environment and the causal consequences of any action are perfectly identical for any agent facing DP(XDT). We can therefore not exclude DP(XDT) on the grounds that it offers different options to different types of decision makers. Moreover, Newcomb's problem requires that it is possible to accurately predict the actions of the decision maker, and one could object to it on this ground. In contrast, all that is required to set up DP(XDT) is that one can compute the output of XDT, and this must presumably be possible if the decision maker is able to use XDT.

Another possible response to this argument is to say that there is a relevant sense in which agents who follow XDT and YDT are not facing the same decision problem when they are faced with DP(XDT), even though they are in identical environments and have access to the same information. We can note that agents following XDT and YDT will compute different values of P($state_i \mid action_j$) in DP(XDT), at least provided that they know which decision theory they themselves follow. The agents will therefore be in different epistemic states when they make their respective decisions, and one could argue that this means that they are not facing the \enquote{same} decision problem in the relevant sense. If this is the case then DP(XDT) fails to demonstrate that there is a decision problem in which XDT is outperformed by another decision theory. 

This response could plausibly be satisfactory, but it calls for a more precise account of what it takes for two decision situations to be \enquote{equivalent} in the relevant sense. There are at least three considerations that seem relevant. First of all, we can say that two decision situations are \enquote{physically equivalent} if they take place in environments with identical dynamics -- that is, the environments have the same states and the same actions, the actions have the same consequences, and corresponding outcomes are associated with the same utilities.\footnote{Stated differently, the situations correspond to identical Markov decision processes (\cite{Bel}).} Furthermore, we can say that two decision situations are \enquote{experientially equivalent} if the agents facing them make identical observations before they make their decisions. Finally, we can say that two decision situations are \enquote{epistemically equivalent} if the decision makers are in the same epistemic state when they make their respective decisions -- that is, they assign the same probability to every proposition.\footnote{I am here of course assuming a broadly Bayesian epistemology, but the argument does not rely much on this assumption.} In most cases physical equivalence would imply experiential equivalence, and experiential equivalence imply epistemic equivalence, but there are cases in which this relationship does not hold. In particular, if an XDT agent and a YDT agent both face DP(XDT) then their situations are physically and experientially equivalent, but not epistemically equivalent. If we consider epistemic equivalence between decision situations to be the most important kind of equivalence then the argument above does not work, since XDT and YDT agents would then not be facing the \enquote{same} decision problem (in the relevant sense) when they are faced with DP(XDT).




It should be noted that this position would also disqualify some Newcomb-like decision problems. For example, say that an evidential decision theorist and a causal decision theorist face Newcomb's problem, and that they know which decision theory they themselves follow. They could then predict which action they will themselves take, from which they could infer what the opaque box contains. This means that they would be in different epistemic states when they make their decisions, and hence not be facing the \enquote{same} decision problem (according to the outlined position). However, if they are not able to predict their own actions then they could be facing the \enquote{same} decision problem. This may be seen as undesirable.

Another possible response to the argument is to say that XDT could be making the \enquote{right} decision in DP(XDT) even though the recommendation of YDT yields more utility than the recommendation of XDT. In situations where the decision maker has incomplete information the rational action may of course be different from the action that yields the greatest amount of utility -- for example, it is not rational to buy a lottery ticket, even if that lottery ticket happens to be a winning ticket. This principle does not apply directly to DP(XDT), since in DP(XDT) the decision maker has information that logically entails the precise state of the environment he is in. However, one could reasonably maintain that XDT should recommend taking the transparent box, because if XDT recommends taking the transparent box then agents following XDT will get \$1,000 in DP(XDT), whereas if it recommends taking the opaque box then agents following XDT will get \$0. If this is the case then DP(XDT) fails to demonstrate that there is any decision problem in which XDT does not make the right decision.

This response could, like the previous response, plausibly be satisfactory. However, it is not entirely unproblematic. First of all, even if we believe that XDT is making the right decision if it takes the transparent box in DP(XDT), we would presumably not want to say that YDT is making the wrong decision if it then takes the opaque box. To make sense of this (without arguing that XDT and YDT are facing different decision problems) it seems as though we would have to argue that the rational course of action in a given situation in some peculiar way depends on what decision theory the decision maker follows. It is not clear to me how exactly this position could be explicated, and I will not attempt to do so here. I will however note that I suspect this approach would end up being effectively equivalent to arguing that XDT and YDT are facing different decision problems.\\

\noindent
\textbf{Acknowledgements}: With many thanks to Caspar Oesterheld for giving feedback on the ideas in this paper.

\printbibliography

@incollection{Nozick1969,
	Author = {Robert Nozick},
	Booktitle = {Essays in Honor of Carl G. Hempel},
	Editor = {Nicholas Rescher et al.},
	Pages = {114-146},
	Publisher = {Springer},
	Title = {Newcomb's Problem and Two Principles of Choice},
	Url = {http://faculty.arts.ubc.ca/rjohns/nozick_newcomb.pdf},
	Year = {1969}
}

@book{Joyce1999-JOYTFO-4,
	publisher = {Cambridge University Press},
	title = {The Foundations of Causal Decision Theory},
	year = {1999},
	author = {James M. Joyce}
}

@book{Jeffrey1965,
	publisher = {University of Chicago Press},
	title = {The Logic of Decision},
	year = {1965},
	author = {Richard C. Jeffrey}
}

@ARTICLE{Bel,
    author = "Richard Bellman",
     title = "A Markovian Decision Process",
   journal = "Indiana Univ. Math. J.",
  fjournal = "Indiana University Mathematics Journal",
    volume = 6,
      year = 1957,
     issue = 4,
     pages = "679--684",
      issn = "0022-2518",
     coden = "IUMJAB",
   mrclass = "",
}

@article{Egan2007,
 ISSN = {00318108, 15581470},
 URL = {http://www.jstor.org/stable/20446939},
 author = {Andy Egan},
 journal = {The Philosophical Review},
 number = {1},
 pages = {93--114},
 publisher = {[Duke University Press, Philosophical Review]},
 title = {Some Counterexamples to Causal Decision Theory},
 volume = {116},
 year = {2007}
}

\end{document}